\documentclass[10pt,twocolumn,letterpaper]{article}

\usepackage{cvpr}              
\definecolor{cvprblue}{rgb}{0.21,0.49,0.74}
\usepackage[pagebackref,breaklinks,colorlinks,allcolors=cvprblue]{hyperref}
\def\paperID{38194} 
\def\confName{CVPR}
\def\confYear{2026}

\usepackage{multirow}
\usepackage{booktabs}

\usepackage{algorithm}
\usepackage{algorithmic}
\usepackage{colortbl}
\usepackage{xurl}
\title{VDE: Training-Free Accelerating Rectified Flow Model via 

Velocity Decomposition and Estimation}

\author{
Junwen Tan \quad 
Jinglin Liang \quad 
Hongyuan Chen \quad 
Shuangping Huang$^*$ \\
South China University of Technology \\
{\tt\small \{eetjw, eeljl, 202230250013\}@mail.scut.edu.cn, eehsp@scut.edu.cn}
\vspace{-0.25in}
}

\newcommand\blfootnote[1]{%
  \begingroup
  \renewcommand\thefootnote{}\footnote{#1}%
  \addtocounter{footnote}{-1}%
  \endgroup
}

\begin{document}
\twocolumn[{%
\renewcommand\twocolumn[1][]{#1}%
\maketitle
\begin{center}
\centering
\includegraphics[width=\linewidth]{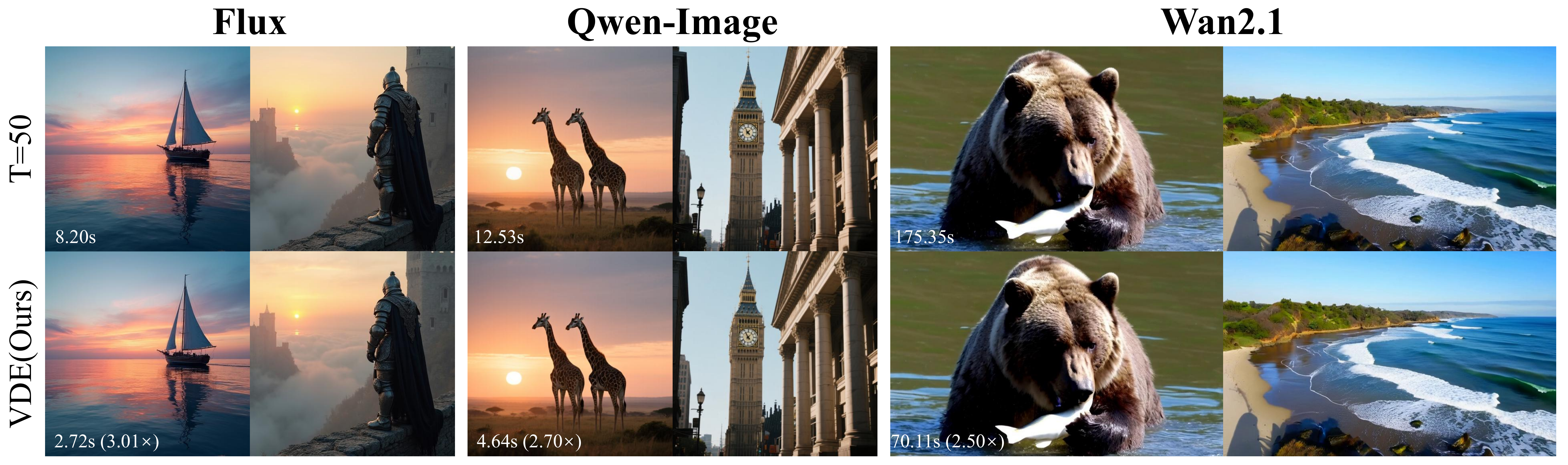}
\captionof{figure}{Qualitative comparison between VDE and standard 50-step sampling across Flux, Qwen-Image, and Wan2.1. VDE achieves almost the same visual quality with significantly reduced runtime.}
\label{fig:cover}
\end{center}
}]

\blfootnote{$^*$ Corresponding author.}

\vspace*{-4mm}
\begin{abstract}
Though rectified flow models have achieved remarkable
performance in image, video, and 3D generation, their
practical deployments are challenged by slow inference speeds. 
Prior acceleration methods reuse cached features from previous steps, which neglects the growing mismatch between static caches and the evolving input, leading to reduced output fidelity.
This work proposes Velocity Decomposition and Estimation (VDE), a training-free acceleration
method that shifts the paradigm from caching-and-reusing to decomposing-and-estimating. 
Specifically, VDE decomposes the model's velocity into components parallel and orthogonal to the input, exploiting their temporal predictability and directional stability for precise, input-adaptive estimation. To prevent error accumulation, it periodically anchors the model’s state via full forward passes.
Extensive experiments on image and video generation tasks demonstrate that VDE achieves substantial acceleration with minimal loss in visual quality. 
Notably, VDE accelerates  Flux by 3.22$\times$ and achieves an LPIPS of 0.069 on Qwen-Image, outperforming the best baseline with a 52.2\% reduction. 
Code: \url{https://github.com/Tan-Junwen/VDE}
\end{abstract}    
\begin{figure*}
    \centering
    \includegraphics[width=1\linewidth]{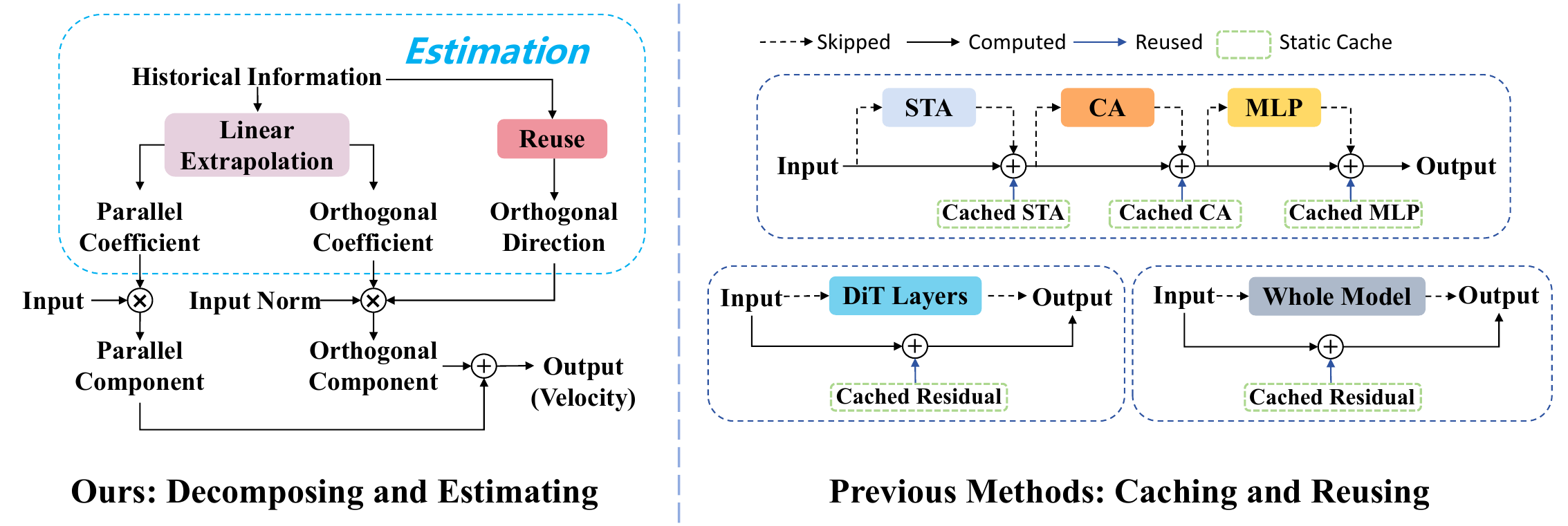}
\caption{
Comparison between standard feature caching and our VDE. \textbf{Left:} VDE decomposes the velocity into parallel and orthogonal components, estimating their evolution via linear extrapolation and directional reuse to ensure adaptive, input-aligned estimation. \textbf{Right:} Prior methods reuse static features, which fail to evolve with the dynamic input and inherently cause cache-input mismatch.
}
\end{figure*}

\section{Introduction}
\label{sec:intro}
Recently, rectified flow models have achieved strong performance in image\cite{flux, sd3-rf, qwenimage}, video\cite{wan, hunyuanvideo, pyramidal}, and 3D asset synthesis\cite{trellis, hunyuan3d, step1x, triposg, craftsman3d}. However, their inference typically requires executing dozens of iterative sampling steps, leading to high latency and limited deployment in real-time or compute-constrained scenarios. 

To mitigate the high inference cost, training-free acceleration methods \cite{deepcache,faster,delta-dit,t-gate,pab} have gained significant traction. 
These approaches typically follow a \textit{cache-and-reuse} paradigm, where intermediate features 
from previous steps are stored and reused in subsequent steps to skip redundant computation. For example, some methods~\cite{pab, omnicache, adacache} cache and reuse the outputs of attention layers, while other methods like TeaCache~\cite{teacache} and EasyCache~\cite{easycache} cache and reuse the residuals of Transformer blocks and the residuals of the whole model, respectively.
Although these methods may trigger reusing based on a low rate-of-change in certain metrics, the fundamental mismatch between the static cache and the dynamic input inevitably propagates into the model's output. This cache-input mismatch results in an output-input mismatch, where the approximated output fails to accurately respond to the current input state, ultimately degrading the visual fidelity of generated content.

In this work, we revisit training-free acceleration and propose  \textbf{Velocity Decomposition and Estimation (VDE)}, a novel method that replaces the static cache reusing with estimation of output directly from the current input via an estimation function, thereby fundamentally eliminating the output-input mismatch caused by cache-input mismatch. This estimation is built upon the discovery of a robust regularity in the model's input-output behavior during sampling. Specifically, we observed that after decomposing the velocity of models into components parallel and orthogonal to the input, there is a stable phase where two consistent patterns appear. First, the scalar coefficients of components evolve smoothly over time, showing strong local linearity between consecutive steps. This property enables reliable estimation of them based on their recent historical values. Second, the orthogonal direction remains largely constant over short intervals, allowing it to be safely reused for multiple adjacent steps without significant error. Based on these insights, VDE constructs an estimation function that dynamically computes the velocity from the current input. Concretely, VDE first executes model inference at periodic anchor steps and decomposes model-predicted velocity to obtain the scalar coefficients and orthogonal direction. Then VDE estimates the coefficients of subsequent steps by linear extrapolation based on their historical values from recent steps and reuse the latest orthogonal direction. The estimated velocity is then synthesized by combining the current input with the extrapolated coefficients and the orthogonal direction. Crucially, unlike caching-based strategies, VDE is inherently input-adaptive---each estimated output is freshly constructed from the current input, eliminating the use of stale features and enabling high-fidelity generation alongside substantial speedups.

Our main contributions are summarized as follows:
\begin{itemize}
    \item We identify the \textbf{cache--input mismatch} problem in existing caching-based acceleration methods and review training-free acceleration as estimating the model's input-output mapping.
    \item We propose \textbf{VDE}, a novel method that shifts the acceleration paradigm from caching-and-reusing to decomposing-and-estimating, enabling precise, input-adaptive estimation of model outputs.
    \item Extensive experiments on image and video generation demonstrate that VDE achieves 2.04-3.22$\times$ speedups with superior visual fidelity, reducing LPIPS by 52.2\% on Qwen-Image against the best baseline.
\end{itemize}


\section{Related Work}
\subsection{Rectified Flow Model}
Diffusion models have dominated high-quality visual generation \cite{ddpm, ddim, glide, imagen-unet3, latent}. Recent transformer-based architectures such as DiT \cite{dit} and related scalable variants \cite{pixart1, pixart2, hunyuandit} further push generation fidelity.
Rectified Flow (RF) models \cite{sd3-rf, fl} reformulate generation as integrating an ODE-driven velocity field, enabling stable training and state-of-the-art synthesis across image \cite{sd3-rf, qwenimage, flux, z-image}, video \cite{wan, hunyuanvideo, pyramidal, opensora}, and 3D \cite{trellis, hi3dgen, hunyuan3d, step1x, craftsman3d, triposg} modalities. Despite their versatility, the sequential nature of RF inference remains computationally demanding, motivating the pursuit of efficient acceleration methods.

\subsection{Training-Free Acceleration Methods.}
Existing acceleration techniques can be broadly grouped into two categories: (1) training-based methods such as distillation \cite{stepzl1, stepzl2, stepzl3, stepzl4}, quantization \cite{he2023ptqd, li2023q, shang2023post, so2023temporal}, consistency training \cite{consistency1, consistency2, consistency3, consistency5} and architecture compression/search \cite{ac1, ac1, ac3},
and (2) training-free methods \cite{deepcache,faster,delta-dit,t-gate}.
 Since training-based methods require additional data and optimization cycles and often reduce generalization ability,  
 training-free methods have emerged as a promising alternative.
Early caching methods for U-Net \cite{unet1, unet2, imagen-unet3} architectures, such as DeepCache\cite{deepcache} and Faster Diffusion\cite{faster}, exploit spatial redundancy by reusing high-level feature maps across nearby timesteps. 
FORA \cite{fora}, $\Delta$-DiT~\cite{delta-dit} and T-GATE~\cite{t-gate} extended this idea to DiT architectures, achieving inference speedup in image generation.
For video generation, PAB~\cite{pab} broadcasts attention outputs in a variance-aware pyramid schedule to maximize reuse. 
AdaCache~\cite{adacache} adjusts feature caching strategies based on content complexity.
TeaCache~\cite{teacache} introduces a timestep-embedding–aware indicator and a rescaling strategy to better predict reuse reliability. EasyCache~\cite{easycache} moves toward runtime adaptivity, dynamically reusing previously computed features without offline profiling. OmniCache~\cite{omnicache} investigates the generation trajectories across samples and uses curvature-guided selection to skip redundant steps.
While effective, these methods reuse stale features, inevitably causing cache-input mismatches that accumulate errors and degrade fidelity.
\begin{figure*}
    \centering
    \includegraphics[width=1\linewidth]{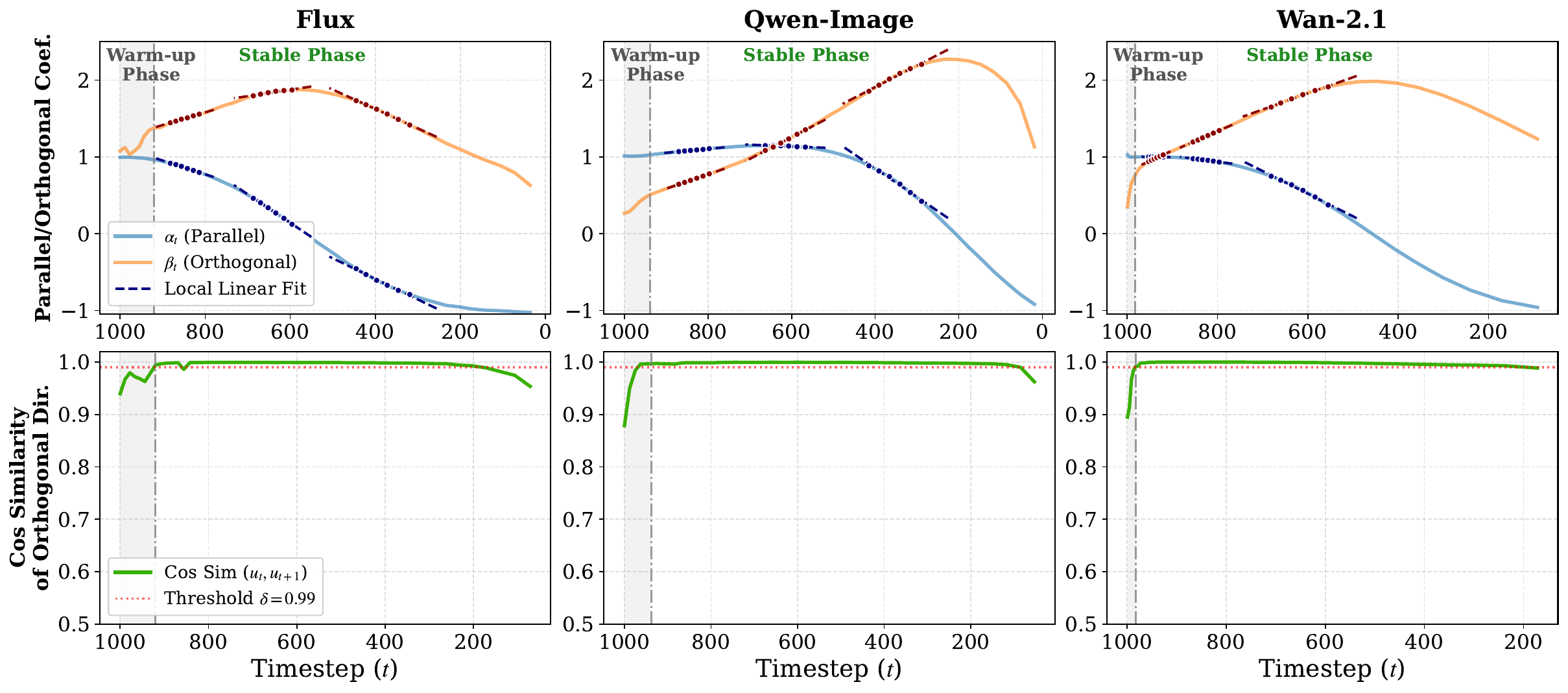}
    \caption{
\textbf{Temporal dynamics of velocity components.} Evolution of the decomposed velocity components across Flux, Qwen-Image, and Wan2.1. \textbf{Top:} The parallel ($\alpha_t$) and orthogonal ($\beta_t$) coefficients. \textbf{Bottom:} The cosine similarity of adjacent orthogonal directions ($\mathbf{u}_t$). After an initial warm-up phase (shaded), the scalar coefficients evolve smoothly with strong local linearity, while the orthogonal direction remains highly stable (cosine similarity ${\approx}1$). These regularities enable VDE's coefficient estimation and direction reuse.
}
\label{fig:observation}
\vspace{-0.1in}
\end{figure*}

\begin{figure}[htbp]  
    \centering
    \includegraphics[width=1\linewidth]{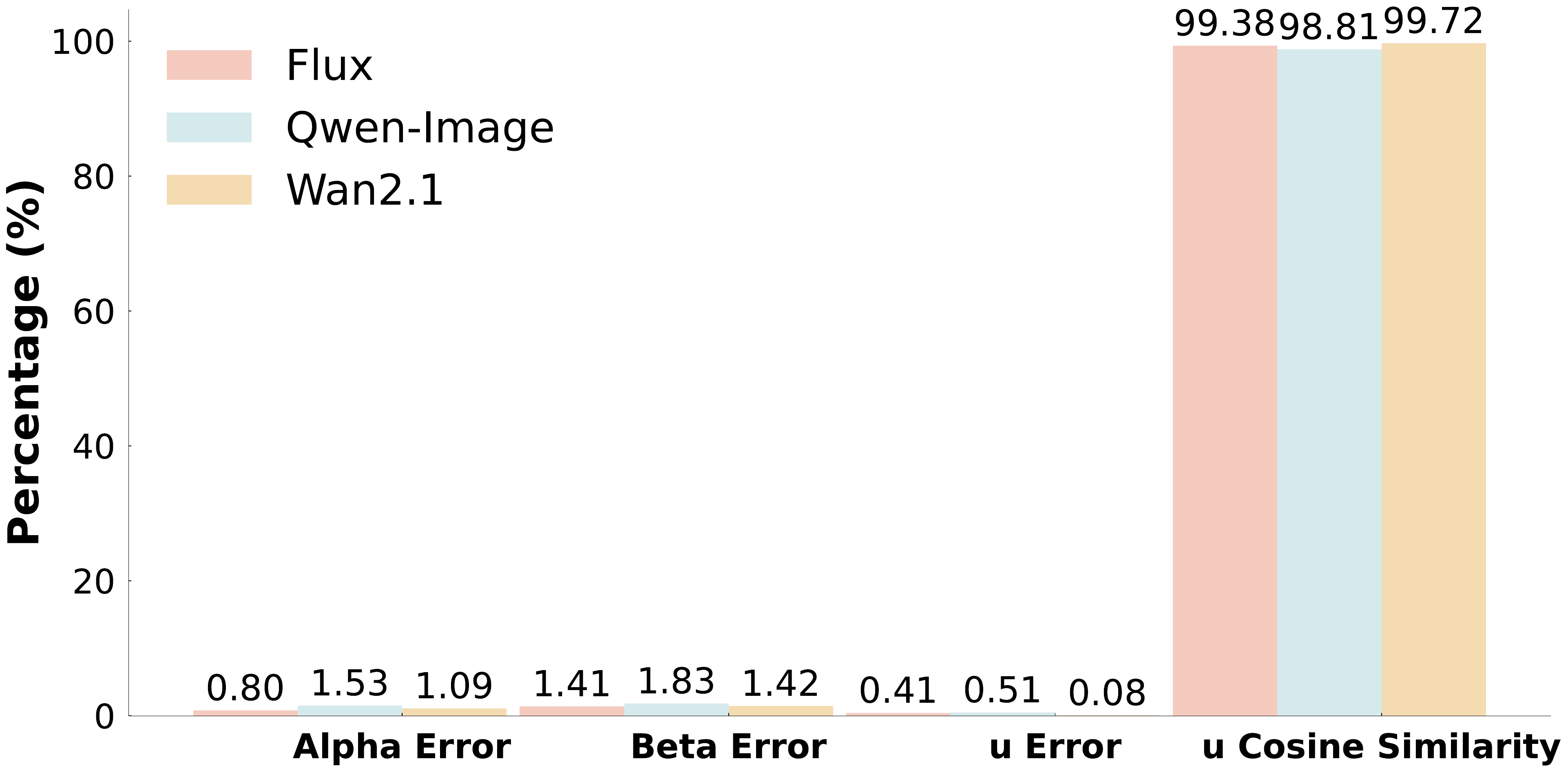}
    \caption{Quantitative evidence for the temporal regularities. We report prediction errors of decomposed velocity components under a two-step linear extrapolation. Both scalar coefficients show extremely low errors, while the orthogonal direction maintains high cosine similarity, confirming their local linearity and short-term stability, respectively.
    }
    \label{fig:quantitative_evidence}  
\vspace{-0.2in}
\end{figure}
\begin{figure*}[t]
    \centering
    \includegraphics[width=1\linewidth]{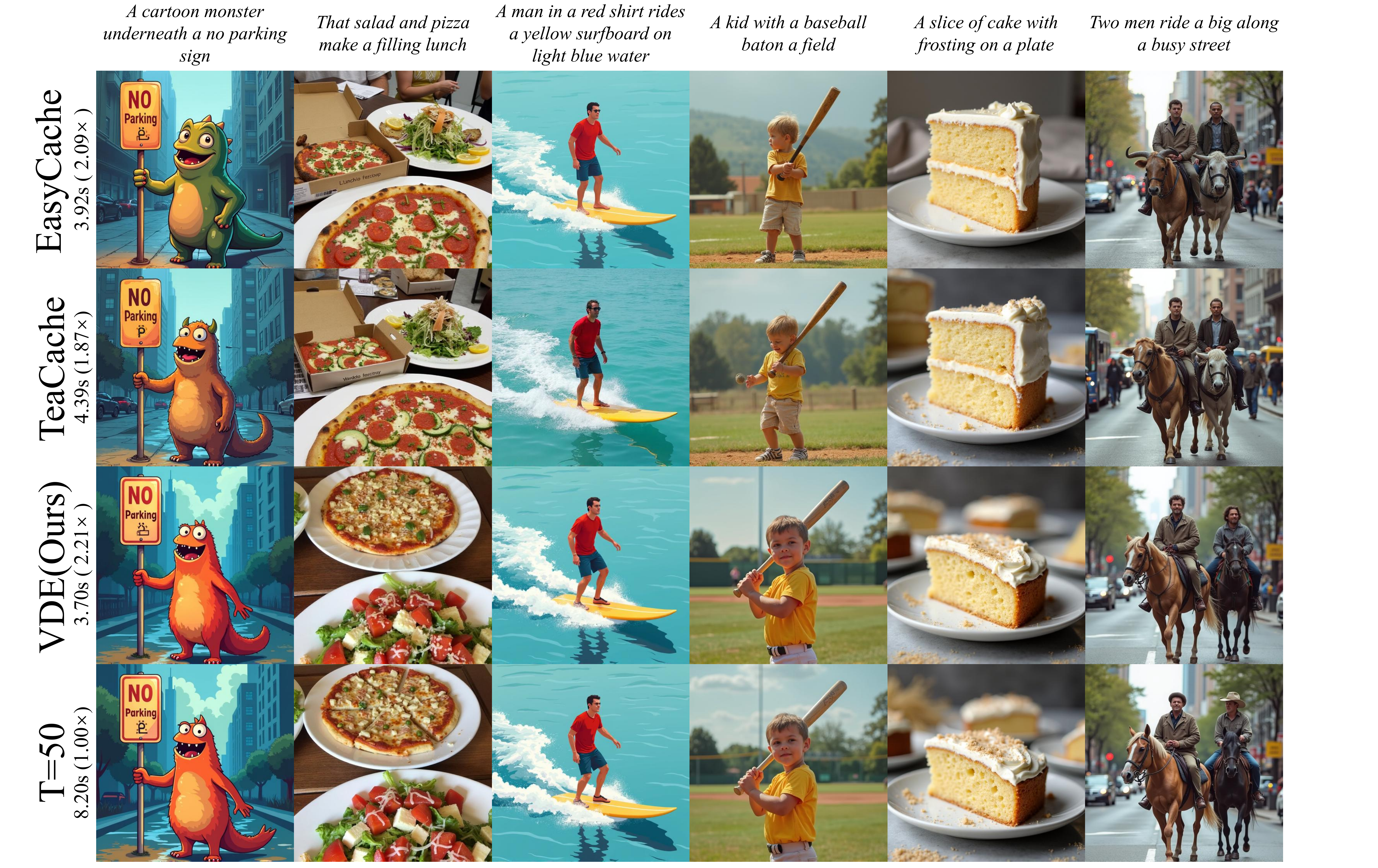}
    \caption{
Qualitative comparison of caching-based methods and our VDE on \text{Flux}. VDE preserves both global structure and fine details.
}
\label{fig:img_visual}
\end{figure*}
\begin{figure*}[t]
    \centering
    \includegraphics[width=1\linewidth]{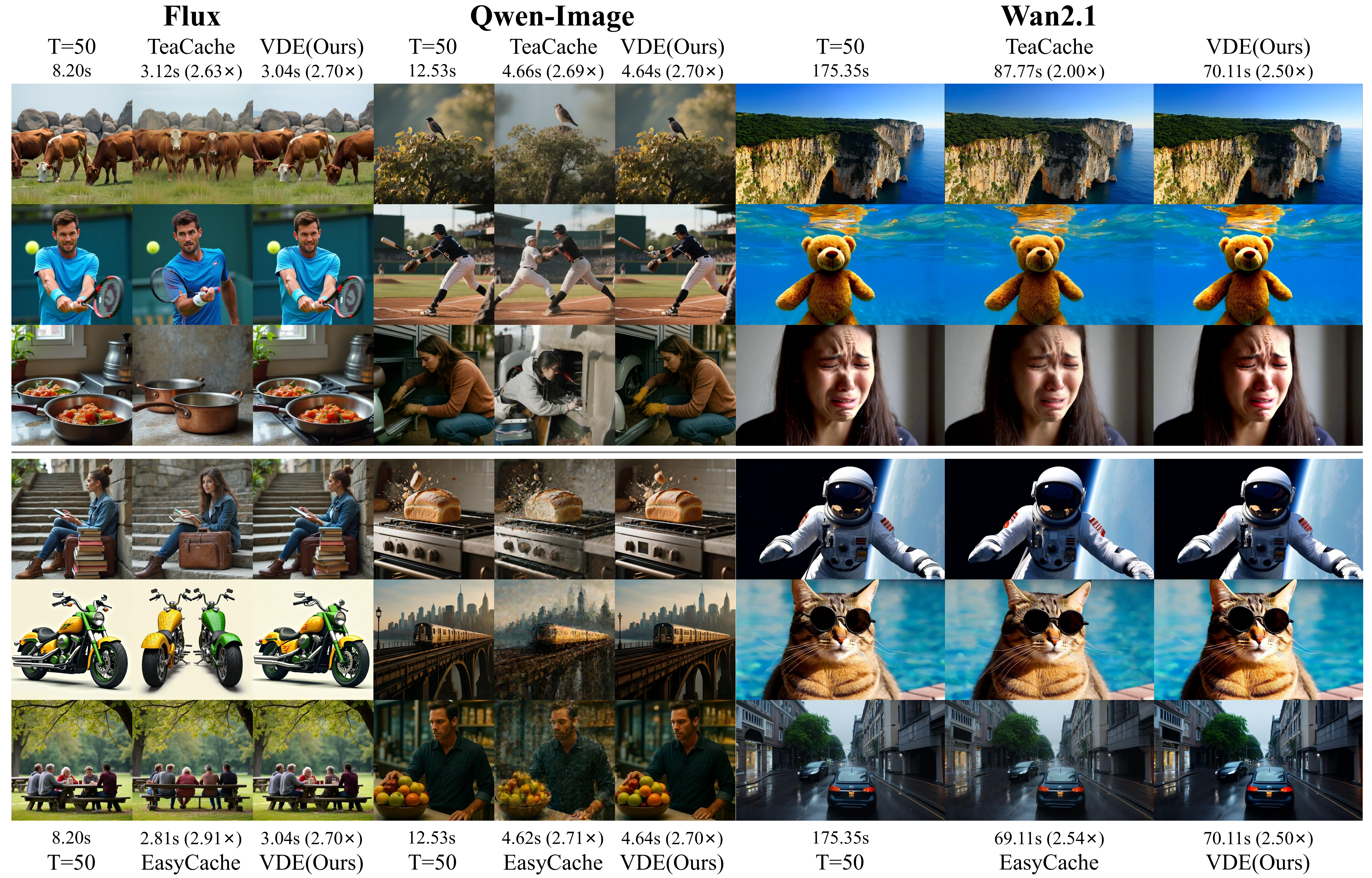}
    \caption{Qualitative comparison of caching-based methods and our VDE on Flux, Qwen-Image, and Wan2.1. 
    VDE successfully accelerates generation while maintaining visual fidelity comparable to the T=50 baseline. In contrast, caching-based methods often exhibit texture degradation or structural inconsistencies due to cache-input mismatch.
    }
    \label{fig:image_visual_2}
\end{figure*}

\section{Methodology}

\subsection{Problem Formulation}
\label{sec:problem}
This work focuses on accelerating Rectified Flow (RF) models \cite{sd3-rf, fl}, which learn a continuous-time generative process to transport data from a noise distribution $p_0$ to a data distribution $p_1$. Given a data sample $x_1$ and Gaussian noise $x_0 \sim \mathcal{N}(0, I)$, an intermediate latent at time $t \in[0, 1]$ is constructed via linear interpolation:
\begin{equation}
    x_t = t x_1 + (1 - t) x_0.
\end{equation}
The core of RF model is to predict a velocity, defined as the time derivative of this path:
\begin{equation}
    \mathbf{v}_t = \frac{dx_t}{dt} = x_1 - x_0.
\end{equation}
During inference, the model $u_\theta$ predicts an instantaneous velocity $v_t = u_\theta(x_t, c, t)$ at each step, which guides the sample $x_t$ from the initial noise $x_0$ towards the final data sample $x_1$ via numerical integration of the ODE $\frac{dx_t}{dt} = v_t$.

We formulate the problem of training-free acceleration for such models as the task of constructing a lightweight estimation function $f_{\text{est}}$. The goal of this function is to accurately approximate the model's velocity output without performing a full forward pass, thereby reducing computational cost. Formally, we aim to obtain:
\begin{equation}
    \hat{v}_t = f_{\text{est}}(x_t, t, I_{\text{hist}}),
\end{equation}
where $\hat{v}_t$ is the estimated velocity, $x_t$ is the current latent input, $t$ is the timestep, and $I_{\text{hist}}$ encapsulates any lightweight historical information from previous steps (e.g., past latents, velocities, or cached features). A well-designed $f_{\text{est}}$ allows us to skip the majority of the model's forward passes while maintaining the visual fidelity of the generated content.

\subsection{Temporal Dynamics of Velocity Components}
\label{sec:observation}
To analyze the temporal behavior of the model, we first introduce an exact decomposition of the velocity output $v_t$ with respect to the input $x_t$. This decomposition forms the basis for our empirical analysis.

\textbf{Velocity Decomposition.} The velocity $v_t$ at any timestep can be uniquely decomposed into a component parallel to the latent $x_t$ and an orthogonal component:
\begin{equation}
    v_t = \alpha_t x_t + \beta_t \|x_t\| u_t,
\label{eq:decomp}
\end{equation}
where $\alpha_t, \beta_t \in \mathbb{R}$ are scalar coefficients and $u_t \in \mathbb{R}^d$ is a unit vector satisfying $u_t^\top x_t = 0$.

Given $v_t$ from a model forward pass, the decomposition is computed as follows. The coefficient of the parallel component is:
\begin{equation}
    \alpha_t = \frac{\langle v_t, x_t \rangle}{\|x_t\|^2}.
\end{equation}
The orthogonal residual is then:
\begin{equation}
    r_t = v_t - \alpha_t x_t.
\end{equation}
Finally, the unit orthogonal direction and its normalized coefficient are obtained by:
\begin{equation}
    u_t = \frac{r_t}{\|r_t\|}, \quad \beta_t = \frac{\|r_t\|}{\|x_t\|}.
\end{equation}
Thus, the tuple $(\alpha_t, \beta_t, u_t)$ provides a complete parameterization of $v_t$ relative to $x_t$.

\textbf{Empirical Findings.} Although the decomposition in Eq.~\ref{eq:decomp} provides an analytical form for the velocity, the tuple $(\alpha_t, \beta_t, \mathbf{u}_t)$ remains unknown before executing the model at timestep $t$. To understand the behavior of these variables, we analyze 500 samples for each model and apply the decomposition across all timesteps for each sample. Our analysis reveals strong and consistent temporal regularities. As shown in Fig.~\ref{fig:observation}, after an initial \textit{warm-up phase}, the sampling trajectory enters a \textit{stable phase} characterized by two patterns: (1) \textbf{Predictable Coefficients.} The scalar coefficients $\alpha_t$ and $\beta_t$ evolve smoothly and demonstrate strong local linearity between consecutive steps. This regularity enables their reliable estimation via extrapolation from recent historical values. (2) \textbf{Stable Orthogonal Direction.} The unit direction $u_t$ remains nearly constant over short intervals, with cosine similarity between adjacent steps typically exceeding 0.99. This permits its safe reuse across multiple steps without introducing significant error.

The quantitative results in Fig.~\ref{fig:quantitative_evidence} robustly support these observations. When employing linear extrapolation during stable phase over an interval of two steps, the coefficients demonstrate remarkably low prediction errors ($\alpha$ error: 0.80\%, 1.53\%, 1.09\%; $\beta$ error: 1.41\%, 1.83\%, 1.42\% across models), confirming the local linearity. Similarly, reusing the orthogonal direction introduces minimal error (0.41\%, 0.51\%, 0.08\%), unequivocally validating its stability. Collectively, these findings imply that, during the stable phase, the velocity $v_t$ can be accurately estimated by analytically propagating its decomposed components, bypassing the need for a full model forward pass at every step.

\textbf{Definition and Identification of the Stable Phase.} Formally, we define that the denoising process enters the stable phase at step $i$ if and only if the linear extrapolation from steps $i$ and $i+1$ accurately predicts step $i+2$, and the orthogonal direction remains stable. This is satisfied when the following criteria are met:
\begin{equation}
    \begin{cases}
        \max\left(\dfrac{|\hat{\alpha}_{i+2} - \alpha_{i+2}|}{|\alpha_{i+2}|}, 
        \dfrac{|\hat{\beta}_{i+2} - \beta_{i+2}|}{|\beta_{i+2}|}\right) < \epsilon \\
        u_i^\top u_{i+1} > \delta
    \end{cases},
\end{equation}
where $\hat{\alpha}_{i+2}$ and $\hat{\beta}_{i+2}$ are the linearly extrapolated coefficients based on steps $i$ and $i+1$. We empirically set the thresholds to $\epsilon=0.02$ and $\delta=0.99$. While this formal formulation enables practical dynamic detection, our extensive profiling across 10,000 trajectories confirms that this transition consistently occurs in the first few steps. To ensure the model reliably establishes the global outline during early denoising steps, we set a conservatively fixed warm-up step in our main experiments. Detailed statistical analysis is provided in the supplementary material.

\subsection{Our VDE Method}
\label{sec:vde}
Building on the above observations, we detail the Velocity Decomposition and Estimation (VDE) method. VDE leverages the predictability of the coefficients and the stability of the orthogonal direction to construct an estimation function. The runtime procedure of VDE is as follows. In the initial \textit{warm-up phase}, we perform a full model inference at every step. Upon entering the \textit{stable phase}, VDE switches to an anchor-and-estimate mode. At periodic anchor steps, a full model inference is executed, and the velocity is decomposed to obtain the ground-truth tuple $(\alpha_t, \beta_t, u_t)$. For all subsequent non-anchor steps until the next anchor, the velocity is estimated without model computation. The estimation for a non-anchor step $t$ is performed by first predicting the scalar coefficients via linear extrapolation from the two most recent anchors at $t_1 > t_2$:
\begin{equation}
    \hat{\alpha}_t = \alpha_{t_1} + \frac{\alpha_{t_2} - \alpha_{t_1}}{t_2 - t_1} (t - t_1), \quad \hat{\beta}_t = \beta_{t_1} + \frac{\beta_{t_2} - \beta_{t_1}}{t_2 - t_1} (t - t_1).
\end{equation}
The orthogonal unit direction is reused from the most recent anchor:
\begin{equation}
    \hat{u}_t = u_{t_2}.
\end{equation}
The final velocity estimate is then synthesized analytically by combining these components with the current input $x_t$:
\begin{equation}
    \hat{v}_t = f_{\text{est}}^{\text{VDE}}(x_t, t, I_{\text{hist}}) = \hat{\alpha}_t x_t + \hat{\beta}_t \|x_t\| \hat{u}_t.
\end{equation}

Explicitly depending on $x_t$, this estimation is inherently input-adaptive and leverages observed temporal regularities.
Decomposition and estimation involves only straightforward arithmetic operations between variables, at a negligible computational cost compared to a full model evaluation. The anchor interval controls the speed-quality trade-off: a larger interval yields higher acceleration but may slightly degrade visual fidelity. Empirically, a two-step interval during the stable phase provides an optimal balance.

\begin{table*}[t]
\centering
\caption{
Quantitative comparison in image generation on FLUX.1 [dev] and Qwen-Image. 
Baseline sampling uses $T{=}50$ steps. 
}
\label{tab:flux_main}
\begin{tabular}{l c c c c c c c c}
\toprule
\multirow{2}{*}{\textbf{Methods}} & \multicolumn{3}{c}{\textbf{Efficiency}} & \multicolumn{3}{c}{\textbf{Visual Retention}} & \multirow{2}{*}{\textbf{CLIP $\uparrow$}} & \multirow{2}{*}{\textbf{ImageReward $\uparrow$}} \\
\cmidrule(r){2-4} \cmidrule(lr){5-7}
& \textbf{Speedup $\uparrow$} & \textbf{Latency (s) $\downarrow$} &
\textbf{NFE $\downarrow$} & \textbf{SSIM $\uparrow$} & \textbf{PSNR $\uparrow$} & \textbf{LPIPS $\downarrow$} & \\
\midrule
\multicolumn{9}{c}{\textbf{FLUX.1 [dev] \cite{flux} (512$\times$512)}} \\
\midrule
$T=50$                            & 1.00$\times$ & 8.20 & 50 & -      & -      & -      & 0.3090 & 0.976 \\
$T=24$                            & 2.06$\times$ & 3.98 & 24 & 0.7512 & 19.74 & 0.2264 & 0.3092 & 0.989 \\
$T=18$                            & 2.71$\times$ & 3.03 & 18 & 0.6862 & 17.79 & 0.3367 & 0.3104 & \textbf{0.991} \\
Teacache-fast \cite{teacache}   & 2.63$\times$ & 3.12 & -  & 0.6858 & 18.07 & 0.3355 & 0.3099 & 0.988 \\
Teacache-slow \cite{teacache}   & 1.87$\times$ & 4.39 & -  & 0.7241 & 19.12 & 0.2941 & 0.3089 & 0.990 \\
Easycache-fast \cite{easycache} & 2.91$\times$ & 2.81 & -  & 0.7240 & 19.59 & 0.3197 & \textbf{0.3109} & 0.986 \\
Easycache-slow \cite{easycache} & 2.09$\times$ & 3.92 & -  & 0.7428 & 19.81 & 0.2793 & 0.3096 & 0.980 \\
\rowcolor{gray!15}
VDE-fast (Ours)                 & 3.01$\times$ & 2.72 & 16  & 0.8267 & 23.19 & 0.1997 & \textbf{0.3109} & 0.969 \\
\rowcolor{gray!15}
VDE-medium (Ours)               & 2.70$\times$ & 3.04 & 18  & 0.8499 & 24.02 & 0.1679 & 0.3102 & 0.973 \\
\rowcolor{gray!15}
VDE-slow (Ours)                 & 2.21$\times$ & 3.70 & 22  & \textbf{0.8877} & \textbf{25.81} & \textbf{0.1243} & 0.3095 & 0.978 \\
\midrule
\multicolumn{9}{c}{\textbf{Qwen-Image \cite{qwenimage} (512$\times$512)}} \\
\midrule
$T=50$                            & 1.00$\times$ & 12.53 & 100 & -      & -      & -      & 0.3156 & \textbf{1.295} \\
$T=24$                            & 2.06$\times$ & 6.09  & 48 & 0.8293 & 22.31 & 0.1744 & 0.3158 & 1.281 \\
$T=18$                            & 2.71$\times$ & 4.62  & 36 & 0.7654 & 19.84 & 0.2403 & 0.3164 & 1.286 \\
Teacache-fast \cite{teacache}   & 2.69$\times$ & 4.66  & -  & 0.5596 & 14.43 & 0.4773 & 0.3116 & 0.948 \\
Teacache-slow \cite{teacache}   & 1.75$\times$ & 7.17  & -  & 0.5573 & 13.99 & 0.4495 & 0.3118 & 1.121 \\
Easycache-fast \cite{easycache} & 2.71$\times$ & 4.62  & -  & 0.7027 & 19.84 & 0.4239 & \textbf{0.3209} & 0.992 \\
Easycache-slow \cite{easycache} & 1.97$\times$ & 6.36  & -  & 0.8708 & 23.83 & 0.1445 & 0.3153 & 1.282 \\
\rowcolor{gray!15}
VDE-fast (Ours)                 & 2.70$\times$ & 4.64  & 36  & 0.8967 & 25.46 & 0.1096 & 0.3163 & 1.287 \\
\rowcolor{gray!15}
VDE-slow (Ours)                 & 2.04$\times$ & 6.14  & 48  & \textbf{0.9362} & \textbf{28.58} & \textbf{0.0691} & 0.3159 & \textbf{1.295} \\
\bottomrule
\end{tabular}
\end{table*}

\section{Experiment}
\subsection{Settings}
\textbf{Base Models and Compared Methods.} We evaluate the performance of our method on various models, including FLUX.1 [dev] \cite{flux} and Qwen-Image \cite{qwenimage} for image generation, and Wan2.1-1.3B \cite{wan} for video generation.
All experiments are conducted on NVIDIA A800 80GB GPUs with Pytorch. 
We compare VDE with representative training-free acceleration baselines, including $\Delta$-DiT\cite{delta-dit}, T-GATE \cite{t-gate}, PAB \cite{pab}, TeaCache \cite{teacache}, and EasyCache \cite{easycache}, and report results of naively reducing steps.

\noindent\textbf{Evaluation Metrics.} To assess the effectiveness of our method, we consider both efficiency and visual quality. 
The efficiency is measured by latency, speedup, and Number of Function Evaluations (NFE). For image generation tasks, we use SSIM, PSNR, and LPIPS \cite{lpips, gong2025monocular} to assess visual retention, human-aligned metrics (e.g., ImageReward~\cite{imagereward}) to validate perceptual fidelity, and CLIP \cite{clip} for semantic alignment.
For video generation tasks, we apply SSIM, PSNR, and LPIPS to assess visual retention, and evaluate performance using the VBench \cite{vbench} framework.

\noindent\textbf{Implementation Details.} All experiments are carried out on NVIDIA A800 80GB GPUs with PyTorch. 
For FLUX.1 [dev], we evaluate three VDE variants: each runs the first 7 steps and the final step, and uses intervals of 2, 3, or 4 steps in between.
For Qwen-Image, we evaluate two VDE variants: each runs the first 11 steps and the final step, and uses intervals of 2 or 5 steps in between.
For Wan2.1, we evaluate two VDE variants: one runs the first 11 steps and the final step with a 2-step interval, and the other runs the first 9 steps and the final step with a 4-step interval.
For image generation tasks, we randomly select 1000 samples from the MS-COCO 2017 \cite{coco2017} validation set, using a resolution of 512×512 for all images.
For video generation, we test our approach based on the VBench framework \cite{vbench}, generating 5 videos for each of 946 benchmark prompts. These videos are comprehensively evaluated across 16 aspects proposed in the VBench evaluation system.

\begin{table*}[t]
\centering
\caption{
Quantitative comparison in video generation on Wan-2.1. 
Baseline sampling uses $T{=}50$ steps. 
}
\label{tab:wan2.1_main}
\begin{tabular}{l c c c c c c c c}
\toprule
\multirow{2}{*}{\textbf{Methods}} & \multicolumn{3}{c}{\textbf{Efficiency}} & \multicolumn{3}{c}{\textbf{Visual Retention}} & \multirow{2}{*}{\textbf{VBench (\%) $\uparrow$}} \\
\cmidrule(r){2-4} \cmidrule(lr){5-7}
 & \textbf{Speedup $\uparrow$} & \textbf{Latency (s) $\downarrow$} &
\textbf{NFE $\downarrow$} & \textbf{SSIM $\uparrow$} & \textbf{PSNR $\uparrow$} & \textbf{LPIPS $\downarrow$} & \\
\midrule
\multicolumn{8}{c}{\textbf{Wan2.1-1.3B \cite{wan} (81 frames, 832$\times$480)}} \\
\midrule
$T=50$                     & $1\times$    & 175.35 & 100 & -      & -      & -      & \textbf{81.30} \\
$T=24$                     & 2.08$\times$ & 84.17  & 48 & 0.5939 & 15.47  & 0.3690 & 80.73 \\
$T=20$                     & 2.50$\times$ & 70.10  & 40 & 0.5226 & 14.50  & 0.4374 & 80.30 \\
Random 0.4                 & 2.45$\times$ & 71.69  & -  & 0.4204 & 11.92  & 0.5911 & 78.68 \\
Static cache               & 2.45$\times$ & 71.54  & -  & 0.5007 & 14.18  & 0.4789 & 79.58 \\
PAB \cite{pab}             & 1.72$\times$ & 102.03 & -  & 0.6484 & 18.84  & 0.3010 & 77.60 \\
TeaCache \cite{teacache}   & 2.00$\times$ & 87.77  & -  & 0.8057 & 22.57  & 0.1277 & 81.04 \\
EasyCache \cite{easycache} & 2.54$\times$ & 69.11  & -  & 0.8337 & 25.24  & 0.0952 & 80.49 \\
\rowcolor{gray!15}
VDE-fast (ours)            & 2.50$\times$ & 70.11  & 40  & 0.8658 & 24.69  & 0.0754 & 80.43 \\
\rowcolor{gray!15}
VDE-slow (ours)            & 2.08$\times$ & 84.18  & 48  & \textbf{0.8902} & \textbf{25.92} & \textbf{0.0554} & 80.32 \\
\bottomrule
\end{tabular}
\end{table*}

\subsection{Main Results}
\label{sec:main_results}
\textbf{Quantitative Results on Image Generation.}
Tab.~\ref{tab:flux_main} reports the efficiency–fidelity trade-off on FLUX.1 [dev] and Qwen-Image.
For reference, we include: (1) the original sampler with $T{=}50$ steps, (2) reduced-step baselines with comparable speedups, and (3) fast/slow variants of TeaCache and EasyCache that respectively favor speed or quality.
We additionally evaluate three variants of VDE (fast/medium/slow) to cover different latency budgets.
Across both base models, VDE consistently achieves a superior speed–quality balance.
On FLUX.1 [dev], \textbf{VDE-fast} matches the latency of EasyCache-fast (3.01$\times$ vs. 2.91$\times$ speedup) by effectively reducing the NFE to 16, yet it delivers substantially higher visual retention, improving SSIM from 0.7240 to 0.8267 and LPIPS from 0.3197 to 0.1997.
\textbf{VDE-medium} further improves reconstruction fidelity with just 18 NFEs while maintaining a competitive 2.70$\times$ speedup.
\textbf{VDE-slow} achieves the highest visual retention overall.
It obtains a 2.21$\times$ speedup while preserving nearly identical visual quality, outperforming the best baseline EasyCache-slow by 19.5\% in SSIM, 30.3\% in PSNR, and reducing LPIPS by 55.4\%. 
Furthermore, VDE exhibits excellent perceptual fidelity on human-aligned evaluations. Notably, VDE-slow achieves an ImageReward of 1.295 on Qwen-Image, which matches the full $T{=}50$ baseline.
This demonstrates that our estimation yields output quality nearly indistinguishable from the original $T{=}50$ baseline while drastically reducing computation.

\noindent\textbf{Quantitative Results on Video Generation.}
We further evaluate VDE on video generation with Wan2.1. As summarized in Table~\ref{tab:wan2.1_main}, VDE achieves competitive speedup (reducing NFEs from 50 to 20-24) while attaining the best visual retention. It outperforms other approaches in both structural and perceptual similarity metrics, and maintains a VBench score close to the original model. These results confirm VDE as an effective training-free accelerator that delivers a superior speed–fidelity trade-off.

\noindent\textbf{Visualization.} Fig.~\ref{fig:img_visual} and Fig.~\ref{fig:image_visual_2} qualitatively compare VDE against training-free baselines (EasyCache, TeaCache) and the original $T{=}50$ model. While EasyCache and TeaCache effectively reduce steps, their outputs deviate noticeably from the $T{=}50$ baseline, exhibiting softened textures, weakened high-frequency details, and compromised fine structures. In contrast, VDE closely matches the $T{=}50$ baseline, preserving global composition and local details (e.g., shapes, textures, shading). This demonstrates that VDE better maintains the generative trajectory and mitigates cache-input mismatches, yielding highly faithful outputs alongside substantial acceleration.

\begin{table}[t]
\centering
\caption{
Ablation on substituting true vs.\ estimated decomposition components on FLUX.1 [dev], 512$\times$512 resolution. The estimated values are obtained on the setting of inference in the first 9 steps and the last step, with an interval of 2 steps in between.
}
\label{tab:flux_kqv}
\setlength{\tabcolsep}{4pt}
\renewcommand{\arraystretch}{1.15}

\begin{tabular*}{\columnwidth}{@{\extracolsep{\fill}} p{2.35cm} c c c c @{}}
\toprule
\textbf{Setting} & \textbf{SSIM $\uparrow$} & \textbf{PSNR $\uparrow$} & \textbf{LPIPS $\downarrow$} & \textbf{CLIP $\uparrow$} \\
\midrule
True $u_t$               & 0.9893 & 40.79 & 0.0132 & 0.3089 \\
True $\beta_t$                     & 0.9262 & 28.76 & 0.0874 & 0.3097 \\
True $\alpha_t$                     & 0.9263 & 28.78 & 0.0860 & 0.3097 \\
True $u_t, \beta_t$            & 0.9916 & 41.79 & 0.0100 & 0.3089 \\
True $u_t, \alpha_t$            & 0.9899 & 41.02 & 0.0124 & 0.3089 \\
True $\alpha_t, \beta_t$                  & 0.9265 & 28.78 & 0.0874 & 0.3098 \\
\textit{Estim.}$\alpha_t, \beta_t, u_t$ & 0.8931 & 26.15 & 0.1198 & 0.3095 \\
\bottomrule
\end{tabular*}
\end{table}

\begin{table}[t]
\centering
\small
\setlength{\tabcolsep}{4pt} 
\caption{Comparison of different anchor intervals on FLUX.1 [dev]. The inference is fixed in the first 7 steps and the last step, with an interval of $n$ steps in between. The resolution is 512$\times$512.}
\label{tab:flux_n}
\begin{tabular*}{\columnwidth}{@{\extracolsep{\fill}} l c c c c c @{}}
\toprule
\multirow{2}{*}{\textbf{Interval}} & \textbf{Efficiency} & \multicolumn{3}{c}{\textbf{Visual Retention}} & \multirow{2}{*}{\textbf{CLIP $\uparrow$}} \\
\cmidrule(r){2-2} \cmidrule(lr){3-5}
& \textbf{Speedup $\uparrow$} & \textbf{SSIM $\uparrow$} & \textbf{PSNR $\uparrow$} & \textbf{LPIPS $\downarrow$} & \\
\midrule
$n=1$ & 1.69$\times$ & 0.9283 & 28.76 & 0.0801 & 0.3093 \\
$n=2$ & 2.21$\times$ & 0.8877 & 25.81 & 0.1243 & 0.3095 \\
$n=3$ & 2.70$\times$ & 0.8499 & 24.02 & 0.1679 & 0.3102 \\
$n=4$ & 3.01$\times$ & 0.8267 & 23.19 & 0.1997 & 0.3109 \\
$n=5$ & 3.22$\times$ & 0.8064 & 22.56 & 0.2168 & 0.3110 \\
\bottomrule
\end{tabular*}
\vspace{-0.05in}
\end{table}

\begin{table}[t]
\centering
\small
\setlength{\tabcolsep}{4pt} 
\caption{Comparison of different resolutions and aspect ratio on Qwen-Image. The inference is fixed in the first 11 steps and the last step, with an interval of $2$ steps in between.}
\label{tab:qwenimage_resolution}
\begin{tabular*}{\columnwidth}{@{\extracolsep{\fill}} l c c c c c @{}}
\toprule
\multirow{2}{*}{\textbf{Resolution}} & \textbf{Efficiency} & \multicolumn{3}{c}{\textbf{Visual Retention}} & \multirow{2}{*}{\textbf{CLIP $\uparrow$}} \\
\cmidrule(r){2-2} \cmidrule(lr){3-5}
& \textbf{Speedup $\uparrow$} & \textbf{SSIM $\uparrow$} & \textbf{PSNR $\uparrow$} & \textbf{LPIPS $\downarrow$} & \\
\midrule
256$\times$256 & 2.07$\times$ & 0.9079 & 26.93 & 0.1036 & 0.3149 \\
512$\times$512 & 2.04$\times$ & 0.9354 & 28.50 & 0.0697 & 0.3159 \\
1024$\times$768 & 2.05$\times$ & 0.9584 & 30.40 & 0.0564 & 0.3150 \\
1024$\times$1024 & 2.06$\times$ & 0.9482 & 30.59 & 0.0541 & 0.3170 \\
\bottomrule
\end{tabular*}
\vspace{-0.1in}
\end{table}

\subsection{Ablation Studies}

\paragraph{Effect of Decomposition Components.}
We first examine how each component in the velocity decomposition affects reconstruction quality. 
Tab.~\ref{tab:flux_kqv} reports results when substituting the true orthogonal direction $u_t$ or the scalar coefficients $(\alpha_t, \beta_t)$ with their estimated counterparts. 
Replacing the true coefficients with our estimates yields nearly lossless reconstruction (e.g., SSIM 0.9893), confirming that the coefficient estimation is highly accurate. 
Replacing the true $u_t$ with the cached direction introduces a moderate quality drop but remains stable, indicating that reusing the orthogonal unit direction is a controlled and low-error approximation. 
When all components are estimated jointly, VDE maintains reasonable reconstruction fidelity, validating that the velocity decomposition can be reliably inferred online without requiring ground-truth intermediate states.

\vspace*{-0.05in}
\paragraph{Anchor Interval.}
We further study the effect of the inference interval $n$, which determines how frequently the velocity decomposition is updated during sampling. 
The anchor interval controls the trade-off between speed and quality. A larger interval yields higher acceleration but may slightly degrade visual fidelity.
Tab.~\ref{tab:flux_n} shows that reducing the update frequency (larger $n$) improves efficiency but gradually decreases visual retention. 
Notably, $n{=}2$ achieves the best speed–quality balance, providing a $2.21\times$ speedup while retaining high fidelity, which corresponds to the VDE-slow configuration reported in the main results. 

\vspace*{-0.05in}
\paragraph{Resolution and Aspect Ratio Generalization.}
Finally, we evaluate VDE under varying output resolutions and aspect ratios on Qwen-Image (Tab.~\ref{tab:qwenimage_resolution}). 
Across 256$\times$256 to 1024$\times$1024, VDE maintains consistent speedups (~2$\times$) and stable visual retention, indicating that our method scales robustly with spatial resolution. 
Importantly, CLIP scores remain nearly unchanged, demonstrating that the semantic alignment is preserved even when image complexity increases. 
These results verify that VDE is resolution-agnostic and generalizes well across modern diffusion model deployment settings.

\section{Conclusion}
We presented Velocity Decomposition and Estimation (VDE), a training-free acceleration framework for rectified flow models. 
This work provides a new perspective by formulating acceleration as estimating the model's output from the current input.
Instead of caching and reusing static features, VDE decomposes the velocity into parallel and orthogonal components and performs lightweight online estimation. 
This shifts the acceleration paradigm from feature reuse to dynamic estimation, effectively mitigating cache–input mismatch and preserving generation fidelity.

\section*{Acknowledgements}
This work was supported by the National Key Research and Development Program of China (No.2023YFC3502900), the National Natural Science Foundation of China (Nos.62176093, 61673182 and 62576139), and the Guangdong Emergency Management Science and Technology Program (No.2025YJKY001).
{
    \small
    \bibliographystyle{ieeenat_fullname}
    \bibliography{main}
}


\end{document}